\documentclass{article}

\usepackage[final, nonatbib]{neurips_2019}
\usepackage[numbers]{natbib}

\usepackage[utf8]{inputenc} 
\usepackage[T1]{fontenc}    
\usepackage{hyperref}       
\usepackage{url}            
\usepackage{booktabs}       
\usepackage{amsfonts}       
\usepackage{nicefrac}       
\usepackage{microtype}      

\usepackage{graphicx}
\usepackage{float}

\usepackage{amssymb}        
\usepackage{pifont}

\usepackage{amsmath}
\usepackage{amssymb}        
\usepackage{algorithm2e}    

\usepackage{xcolor}
\usepackage{booktabs}
\usepackage{pgfplots}

\usepackage{caption}
\usepackage{subcaption}

\usepackage{wrapfig}

\definecolor{mina-green}{rgb}{0.1647, 0.4196, 0.0964}
\definecolor{mina-yellow}{rgb}{0.9999, 0.6226, 0.0000}

\newcommand{\figref}[1]{Figure~\ref{fig:#1}}
\newcommand{\tabref}[1]{Table~\ref{tab:#1}}

\newcommand{\tb}[1]{\textbf{#1}}
\newcommand{\ti}[1]{\textit{#1}}

\newcommand{\textft}[1]{{\fontfamily{lmss}\selectfont{#1}}}

\def\shift{$<$shift$>$}

\def\uniform{\textft{Unif}}
\def\naive{\textft{Linear}}
\def\stopword{\textft{Stopword}}
\def\constrain{\textft{Constr}}

\def\z{z}
\def\x{x}

\def\eps{\epsilon}

\def\mininize{\min_{\qparam, \pparam}}

\def\qparam{\alpha}
\def\pparam{\beta}

\def\q{q_{\qparam}}
\def\p{p_{\pparam}}
\def\pxz{\p(\x\mid\z)}
\def\qzx{\q(\z\mid\x)}

\def\loss{\mathrm{loss}(x, \qparam, \pparam)}
\def\cost{\mathrm{cost}(x, \qparam)}
\def\R{G(\x, \z)}

\def\nablaq{\nabla_{\qparam}}

\def\expec{\mathbb{E}}

\def\expecqzx{\expec_{\q(\z\mid \x)}}

\def\numtokens{\mathrm{\# tokens}(\z)}

\title{Learning Autocomplete Systems \\ as a Communication Game}

\author{
  Mina Lee\textsuperscript{1}, Tatsunori B. Hashimoto\textsuperscript{1,2}, Percy Liang\textsuperscript{1} \\
  \\
  \textsuperscript{1} Department of Computer Science \quad \textsuperscript{2} Department of Statistics \\
  Stanford University \\
  \texttt{\{minalee, pliang\}@cs.stanford.edu} \quad \texttt{thashim@stanford.edu}
}

\begin{document}

\maketitle

\begin{abstract}
We study textual autocomplete---the task of predicting a full sentence from a partial sentence---as a human-machine communication game. 
Specifically, we consider three competing goals for effective communication: use as few tokens as possible (efficiency), transmit sentences faithfully (accuracy), and be learnable to humans (interpretability). 
We propose an unsupervised approach which tackles all three desiderata by constraining the communication scheme to \ti{keywords} extracted from a source sentence for interpretability and optimizing the efficiency-accuracy tradeoff.
Our experiments show that this approach results in an autocomplete system that is 52\% more accurate at a given efficiency level compared to baselines, is robust to user variations, and saves time by nearly 50\% compared to typing full sentences.
\end{abstract}

\section{Introduction}

\begin{wrapfigure}{r}{0.5\textwidth}
    \vspace{-1.25em}
	\centering
	\captionsetup{margin=0.3cm}
    \includegraphics[width=0.43\textwidth]{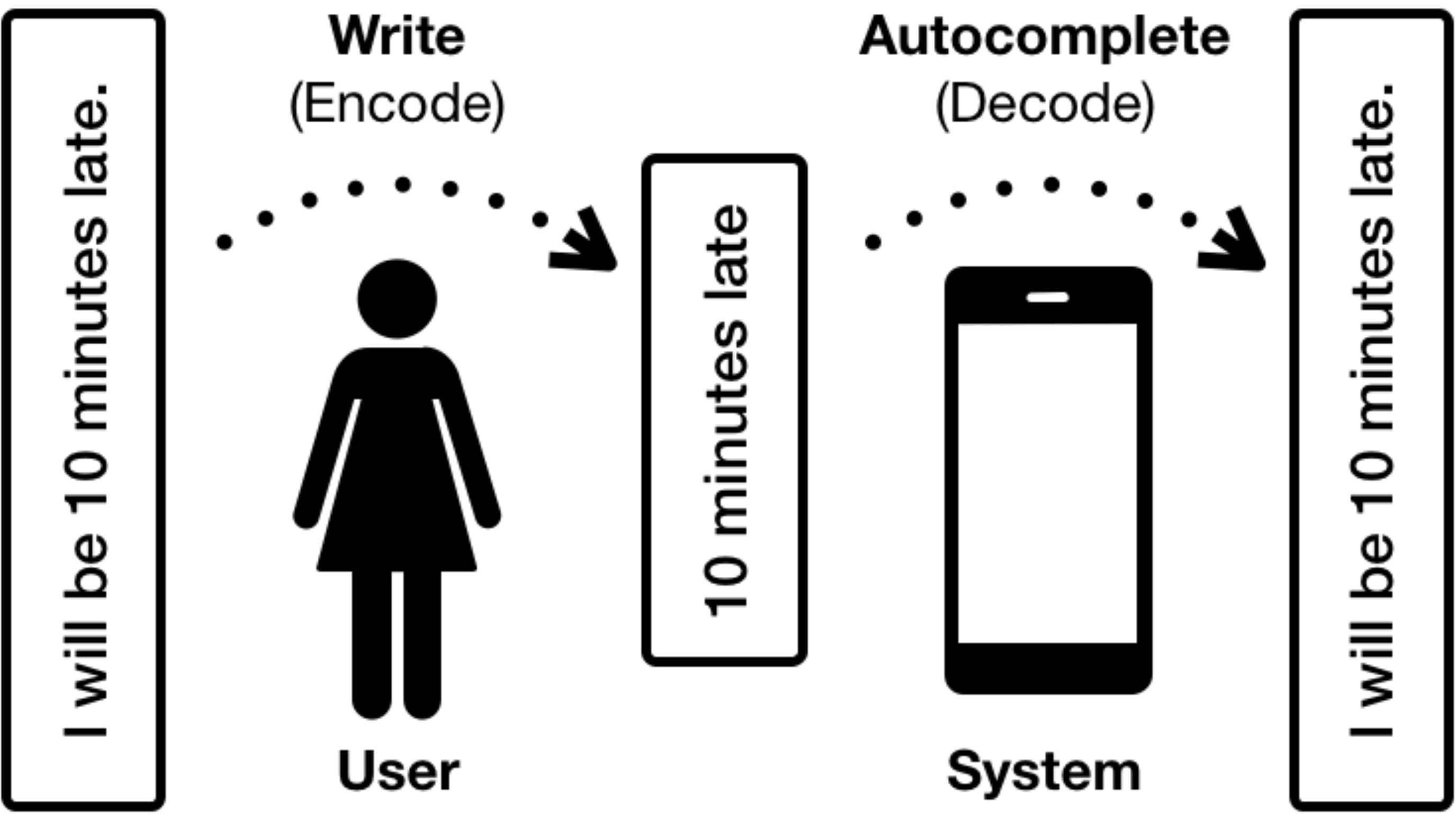}
	\caption{\label{fig:problem}
  We consider a communication game in which a user encodes a sentence into a sequence of keywords, and the system seeks to decode the sentence given only the keywords.
  }
    \vspace{-1.25em}
\end{wrapfigure}

Suppose a user wants to write a sentence ``I will be 10 minutes late.''  Ideally, she would type just a few keywords such as ``10 minutes late'' and an autocomplete system would be able to infer the intended sentence (Figure~\ref{fig:problem}).
Existing left-to-right autocomplete systems~\cite{yossef2011context, svyatkovskiy2019pythia} can often be inefficient, as the prefix of a sentence (e.g. ``I will be'') fails to capture the core meaning of the sentence. Besides the practical goal of building a better \emph{autocomplete} system,
we are interested in exploring the tradeoffs inherent to such communication schemes between the efficiency of typing keywords, accuracy of reconstruction, and interpretability of keywords.

One approach to learn such schemes is to collect a supervised dataset of keywords-sentence pairs as a training set,
but (i) it would be expensive to collect such data from users, and
(ii) a static dataset would not capture a real user's natural predilection to adapt to the system \citep{wang2016games}.
Another approach is to avoid supervision and jointly learn a user-system communication scheme to directly optimize the combination of efficiency and accuracy. However, learning in this way can lead to communication schemes that are uninterpretable to humans \cite{lewis2017deal,mordatch2018emergence} (see Appendix~\ref{sec:appendix_related} for additional related work).

In this work, we propose a simple, unsupervised approach to an autocomplete system that is efficient, accurate, and interpretable.
For interpretability, we restrict keywords to be subsequences of their source sentences based on the intuition that humans can infer most of the original meaning from a few keywords.
We then apply multi-objective optimization approaches to directly control and achieve desirable tradeoffs between efficiency and accuracy.

We observe that naively optimizing a linear combination of efficiency and accuracy terms is unstable and leads to suboptimal schemes.
Thus, we propose a new objective which optimizes for communication efficiency under an accuracy constraint. We show this new objective is more stable and efficient than the linear objective at all accuracy levels.

As a proof-of-concept, we build an autocomplete system within this framework which allows a user to write sentences by specifying keywords. We empirically show that our framework produces communication schemes that are 52.16\% more accurate than rule-based baselines when specifying 77.37\% of sentences, and 11.73\% more accurate than a naive, weighted optimization approach when specifying 53.38\% of sentences.
Finally, we demonstrate that humans can easily adapt to the keyword-based autocomplete system and save nearly 50\% of time compared to typing a full sentence in our user study.

\section{Approach}
\label{sec:approach}

Consider a communication game in which the goal is for a user to communicate a
\emph{target} sequence $\x = (\x_1, ..., \x_m)$ to a system by passing a
sequence of \emph{keywords} $\z = (\z_1, ..., \z_n)$.
The user generates keywords $\z$ using an encoding strategy $\qzx$,
and the system attempts to guess the target sequence $\x$ via a decoding strategy $\pxz$.

A good communication \emph{scheme} $(\q, \p)$ should be both \emph{efficient} and \emph{accurate}.
Specifically, we prefer schemes that use fewer keywords (cost), and the target sentence $\x$ to be reconstructed with high probability (loss) where
\begin{align} 
    \cost &= \expecqzx[\numtokens] \label{eq:cost} \\
    \loss &= \expecqzx[- \log \pxz] \label{eq:loss}.
\end{align}

Based on our assumption that humans have an intuitive sense of retaining important keywords, we restrict the set of schemes to be a (potentially noncontiguous)  \ti{subsequence} of the target sentence.  Our hypothesis is that such subsequence schemes naturally ensure interpretability, as efficient human and machine communication schemes are both likely to involve keeping important content words.

\paragraph{Modeling with autoencoders.}
To learn communication schemes without supervision, we model the cooperative communication between a user and system through an encoder-decoder framework. Concretely, we model the user's encoding strategy $\qzx$ with an encoder which encodes the target sentence $\x$ into the keywords $\z$ by keeping a subset of the tokens. This stochastic encoder $\qzx$ is defined by a model which returns the probability of each token retained in the final subsequence $z$. Then, we sample from Bernoulli distributions according to these probabilities to either keep or drop the tokens independently (see Appendix~\ref{sec:appendix_encoder} for an example). 

We model the autocomplete system's decoding strategy $\pxz$ as a probabilistic model which conditions on the keywords $\z$ and returns a distribution over predictions $\x$. We use a standard sequence-to-sequence model with attention and copying for the decoder, but any model architecture can be used (see Appendix~\ref{sec:appendix_experiments} for details).

\paragraph{Multi-objective optimization.}
Our goal now is to learn encoder-decoder pairs which optimally balance the communication cost and reconstruction loss. The simplest approach to balancing efficiency and accuracy is to weight $\cost$ and $\loss$ linearly using a weight $\lambda$ as follows,
\begin{gather} \label{eq:naive}
    \mininize \expec[\cost] + \lambda \expec[\loss],
\end{gather}
where the expectation is taken over the population distribution of source sentences $x$, which is omitted to simplify notation.
However, we observe that naively weighting and searching over $\lambda$ is suboptimal and highly unstable---even slight changes to the weighting results in degenerate schemes which keep all or none of its tokens. This instability motivates us to develop a new stable objective.

Our main technical contribution is to draw inspiration from the multi-objective optimization literature and view the tradeoff as a sequence of constrained optimization problems, where we minimize the expected cost subject to varying expected reconstruction error constraints $\epsilon$,
\begin{equation}
\underset{\qparam, \pparam}{\text{min}}~\expec[\cost] \quad \text{subject to}~\expec[\loss] \leq \epsilon.
\label{eq:consopt}
\end{equation}
This greatly improves the stability of the training procedure. We empirically observe that the model initially keeps most of the tokens to meet the constraints, and slowly learns to drop uninformative words from the keywords to minimize the cost. Furthermore, $\eps$ in Eq~\eqref{eq:consopt} allows us to directly control the maximum reconstruction error of resulting schemes, whereas $\lambda$ in Eq~\eqref{eq:naive} is not directly related to any of our desiderata.

To optimize the constrained objective, we consider the Lagrangian of Eq~\eqref{eq:consopt},
\begin{equation} \label{eq:constrained}
  \min_{\qparam, \pparam} \max_{\lambda \geq 0} J(\qparam, \pparam, \lambda) \quad \text{where}~J(\qparam, \pparam, \lambda) = \expec[\cost] + \lambda(\expec[\loss] - \eps).
\end{equation}

Much like the objective in Eq~\eqref{eq:naive} we can compute unbiased gradients by replacing the expectations with their averages over random minibatches. Although gradient descent guarantees convergence on Eq~\eqref{eq:constrained} only when the objective is convex, we find that not only is the optimization stable, the resulting solution achieves better performance than the weighting approach in Eq~\eqref{eq:naive}.

\paragraph{Optimization.}
Optimization with respect to $\qzx$ is challenging as $\z$ is discrete, and thus, we cannot differentiate $\qparam$ through $\z$ via the chain rule. Because of this, we use the stochastic REINFORCE estimate~\cite{williams1992simple} as follows:
\begin{gather}\label{eq:pg}
  \nablaq J(\qparam,\pparam,\lambda) = \expec [ \expecqzx[\nablaq \log\qzx \cdot \R] ]\\
  \R = \underbrace{\numtokens}_\text{per-example cost} + \lambda (\underbrace{- \log\pxz}_\text{per-example loss} - \eps).
\end{gather}
We perform joint updates on $(\qparam, \pparam, \lambda)$, where $\pparam$ and $\lambda$ are updated via standard gradient computations, while $\qparam$ uses an unbiased, stochastic gradient estimate where we approximate the expectation in Eq~\eqref{eq:pg}. We use a single sample from $\qzx$ and moving-average of rewards as a baseline to reduce variance.

\section{Experiments}
We evaluate our approach by training an autocomplete system on 500K randomly sampled sentences from Yelp reviews~\cite{yelp2017yelp} (see Appendix \ref{sec:appendix_experiments} for details). 
We quantify the \emph{efficiency} of a communication scheme $(\q,\p)$ by the retention rate of tokens, which is measured as the fraction of tokens that are kept in the keywords. 
The \emph{accuracy} of a scheme is measured as the fraction of sentences generated by greedily decoding the model that \textit{exactly matches} the target sentence.

\begin{figure}[t!]
\centering
\vspace{-1em}
\begin{minipage}{.45\textwidth}
  \centering
  \includegraphics[width=1\linewidth]{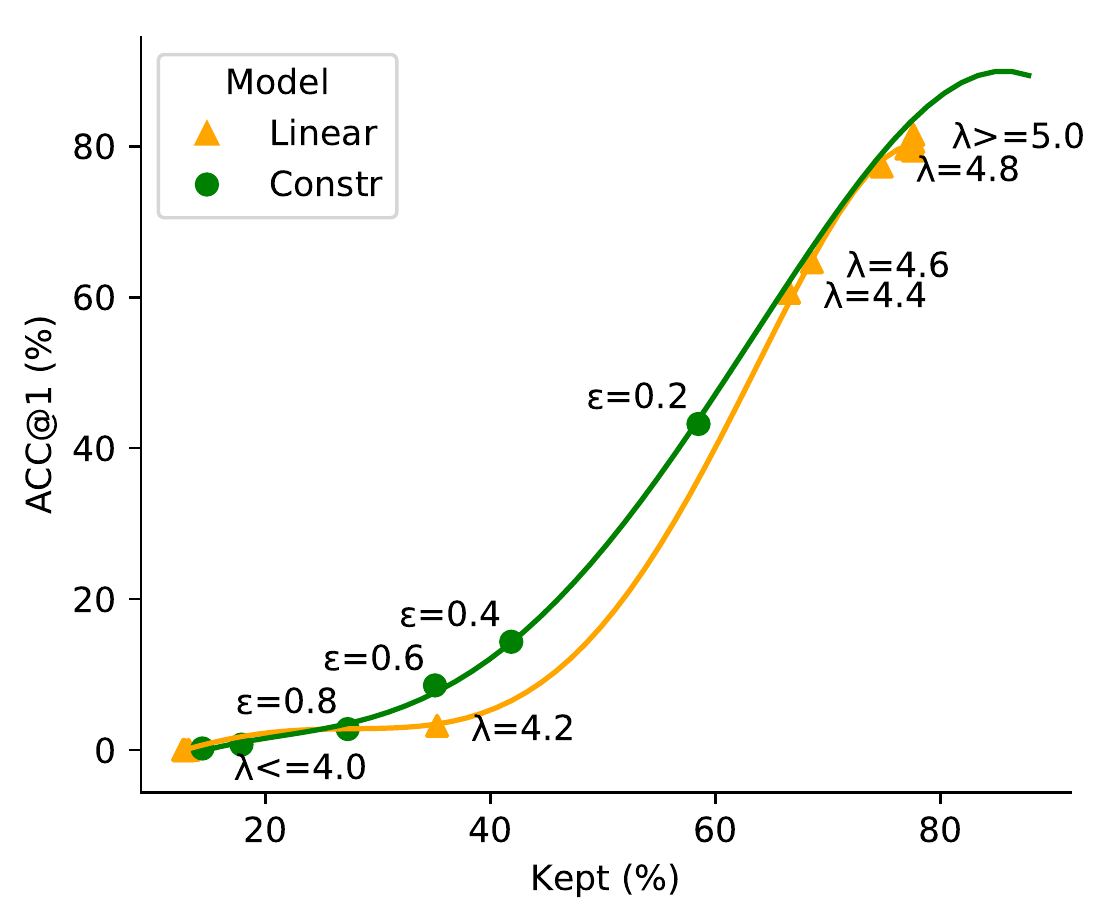}
  \captionof{figure}{Our constrained objective ({\color{mina-green}\constrain}) learns more efficient and accurate schemes than the unconstrained objective ({\color{mina-yellow}\naive}).}
  \label{fig:naiveconstrained}
\end{minipage}
\hfill
\begin{minipage}{.45\textwidth}
  \centering
  \includegraphics[width=1\linewidth]{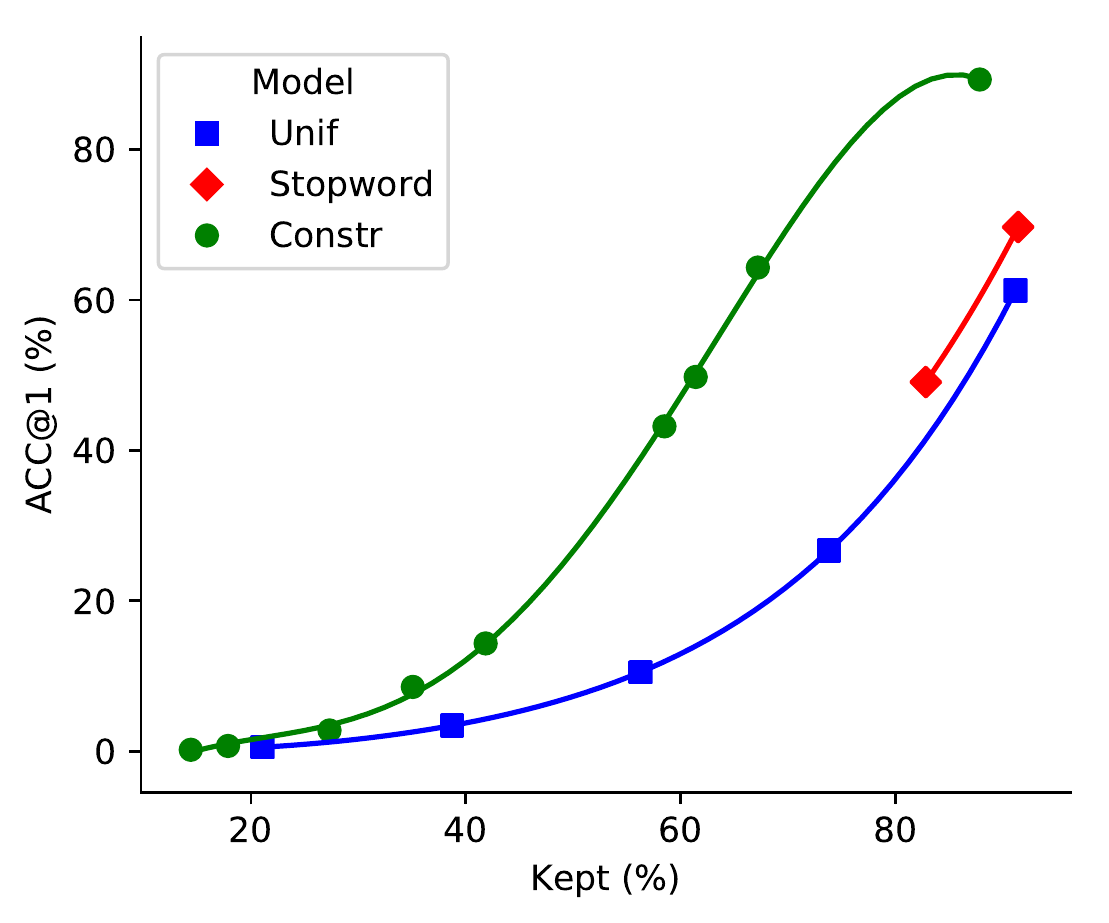}
  \captionof{figure}{Our models ({\color{mina-green}\constrain}) improve the tradeoff between accuracy and efficiency compared to the baselines ({\color{blue}\uniform}~and~{\color{red}\stopword}).}
  \label{fig:performance}
\end{minipage}
\vspace{-0.5em}
\end{figure}

\paragraph{Effectiveness of constrained objective.}
We first show that the linear objective in Eq~\eqref{eq:naive} is suboptimal compared to the constrained objective in Eq~\eqref{eq:consopt}. \figref{naiveconstrained} compares the achievable accuracy and efficiency tradeoffs for the two objectives, which shows that the constrained objective results in more efficient schemes than the linear objective at every accuracy level
(e.g. 11.73\% more accurate at a 53.38\% retention rate). 

We also observe that the linear objective is highly unstable as a function of the tradeoff parameter $\lambda$ and requires careful tuning. Even slight changes to $\lambda$ results in degenerate schemes that keep all or none of the tokens (e.g. $\lambda \leq 4.2$ and $\lambda \geq 4.4$). On the other hand, the constrained objective is substantially more stable as a function of $\eps$ (e.g. points for $\eps$ are more evenly spaced than $\lambda$).

\paragraph{Efficiency-accuracy tradeoff.} 
We quantify the efficiency-accuracy tradeoff compared to two rule-based baselines: \uniform~and \stopword. The \uniform~encoder randomly keeps tokens to generate keywords with the probability $\delta$. The \stopword~encoder keeps all tokens but drops stop words (e.g. `the', `a', `or') all the time ($\delta=0$) or half of the time ($\delta=0.5$). The corresponding decoders for these encoders are optimized using gradient descent to minimize the reconstruction error (i.e. $\loss$). 

\figref{performance} shows that two baselines achieve similar tradeoff curves, while the constrained model achieves a substantial 52.16\% improvement in accuracy at a 77.37\% retention rate compared to \uniform, thereby showing the benefits of jointly training the encoder and decoder.

\begin{table*}[t!]
  \begin{center}
  \vspace{-1.5em}
  \fontsize{8}{10}\selectfont
    \setlength{\tabcolsep}{2.5pt}
    \begin{tabular}{c|ccc}
      \toprule
\tb{Sentence}      & \multicolumn{3}{c}{\textft{She's absolutely wonderful.}}\\
      \cmidrule(lr){1-4}
\tb{Keywords}    & \multicolumn{1}{c}{\textft{wonderful}}
            & \multicolumn{1}{c}{\textft{she wonderful}}
            & \multicolumn{1}{c}{\textft{she's absolutely wonderful}} \\
            
            & \multicolumn{1}{c}{(\textcolor{red}{under-specified})}
            & \multicolumn{1}{c}{}
            & \multicolumn{1}{c}{(\textcolor{red}{over-specified})} \\
      \cmidrule(lr){1-4}    

\tb{Suggestions} & \textft{The entire experience was wonderful.} 
            & \textft{\textcolor{mina-green}{She's absolutely wonderful.}}
            & \textft{\textcolor{mina-green}{She's absolutely wonderful.}} \\
            & \textft{Great service and wonderful food.} 
            & \textft{She's simply wonderful.}
            & \textft{And she's absolutely wonderful} \\ 
            & \textft{Great food and wonderful service.}
            & \textft{She's a wonderful doctor.} 
            & \textft{and she's absolutely wonderful} \\

      \midrule
\tb{Sentence}      & \multicolumn{3}{c}{\textft{Great find for casual dining!}}\\
      \cmidrule(lr){1-4}
\tb{Keywords}    & \multicolumn{1}{c}{\textft{great dining}}
            & \multicolumn{1}{c}{\textft{find casual dining}}
            & \multicolumn{1}{c}{\textft{great find for casual dining}} \\

            & \multicolumn{1}{c}{(\textcolor{red}{under-specified})}
            & \multicolumn{1}{c}{}
            & \multicolumn{1}{c}{(\textcolor{red}{over-specified})} \\
      \cmidrule(lr){1-4}
\tb{Suggestions} & \textft{What a great dining experience.}
            & \textft{\textcolor{mina-green}{Great find for casual dining.}}
            & \textft{\textcolor{mina-green}{Great find for casual dining.}} \\
            & \textft{Overall a great dining experience.}
            & \textft{Perfect find for casual dining.}
            & \textft{Great find for casual dining...} \\
            & \textft{A great dining experience.} 
            & \textft{Great find for casual dining...}  
            & \textft{Great find in casual dining.}  \\
            
      \bottomrule
    \end{tabular}
    \caption{Examples from the autocomplete task. The three columns contain keywords provided by different users and corresponding top three suggested sentences generated by our model.}
    \label{tab:examples}
    \vspace{-1.5em}
  \end{center}
\end{table*}

\paragraph{Robustness and analysis.} We provide additional experimental results on the robustness of learned communication schemes as well as in-depth analysis on the correlation between the retention rates of tokens and their properties, which we defer to Appendix~\ref{sec:appendix_robustness} and~\ref{sec:appendix_analysis} for space.

\paragraph{User study.}
We recruited 100 crowdworkers on Amazon Mechanical Turk (AMT) and measured completion times and accuracies for typing randomly sampled sentences from the Yelp corpus.
Each user was shown alternating autocomplete and writing tasks across $50$ sentences (see Appendix~\ref{sec:appendix_ui} for user interface). For the autocomplete task, we gave users a target sentence and asked them to type a set of keywords into the system. The users were shown the top three suggestions from the autocomplete system, and were asked to mark whether each of these three suggestions was semantically equivalent to the target sentence. For the writing task, we gave users a target sentence and asked them to either type the sentence verbatim or a sentence that preserves the meaning of the target sentence.

Table~\ref{tab:examples} shows two examples of the autocomplete task and actual user-provided keywords. Each column contains a set of keywords and its corresponding top three suggestions generated by the autocomplete system with beam search. We observe that the system is likely to propose generic sentences for under-specified keywords (left column) and almost the same sentences for over-specified keywords (right column). For properly specified keywords (middle column), the system completes sentences accordingly by adding a verb, adverb, adjective, preposition, capitalization, and punctuation.

Overall, the autocomplete system achieved high accuracy in reconstructing the keywords. Users marked the top suggestion from the autocomplete system to be semantically equivalent to the target $80.6$\% of the time, and one of the top 3 was semantically equivalent $90.11$\% of the time. The model also achieved a high exact match accuracy of 18.39\%.
Furthermore, the system was efficient, as users spent $3.86$ seconds typing keywords compared to $5.76$ seconds for full sentences on average. 
The variance of the typing time was $0.08$ second for keywords and $0.12$ second for full sentences, indicating that choosing and typing keywords for the system did not incur much overhead.

\subsubsection*{Acknowledgments}
We thank the reviewers and Yunseok Jang for their insightful comments. This work was supported by NSF CAREER Award IIS-1552635 and an Intuit Research Award.

\subsubsection*{Reproducibility}
All code, data and experiments are available on CodaLab at \url{https://bit.ly/353fbyn}.

\bibliographystyle{plainnat}
\bibliography{all}

\newpage

\appendix
\section{Related work}
\label{sec:appendix_related}

Our goal of learning efficient communication schemes in an autocomplete task is closely related to a rich literature on communication and compression.

\paragraph{Learning communication schemes.} 
Tradeoffs between communication efficiency (both information theoretically and computationally) and accuracy have been studied both in the machine learning \cite{levy2007speakers} and cognitive science \cite{levy2018communicative} literatures, where much of the focus has been on understanding the characteristics of efficient human communication. Our work differs in that we are learning schemes based on objective efficiency and accuracy metrics, while taking human interpretability into account.

Approaches to learning communication schemes based on rational speech acts (RSA) combine intuitions on human communication efficiency (i.e. pragmatics) with an explicit computational algorithm for generating communication schemes \cite{golland2010pragmatics,frank2012pragmatics, smith2013pragmatics, monroe2015pragmatics, goodman2016pragmatic, khani2018pip}. RSA differs from our approach in that it requires a base listener or speaker, which ensures that the resulting schemes are interpretable. In contrast, we constrain our schemes to be keywords and optimize the resulting schemes.

There is a growing literature on learning communication schemes through two-player games. Such problems have been explored in negotiation \cite{lewis2017deal}, cooperative communication \cite{das2017learning}, and multi-agent games \cite{mordatch2018emergence}. The majority of these approaches optimize a single, task-oriented metric (i.e. accuracy). In contrast, our work explicitly considers the tradeoff between task efficiency and accuracy, and shows that using sufficiently restricted schemes results in a simple algorithm for making effective tradeoffs.

\paragraph{Sentence compression and summarization.} 
Our task can be seen as a restricted form of sentence compression or extractive summarization, where the compressed form is a sequence of keywords extracted from its original sentence. 

Most work on sentence compression aims to find a short, grammatical summary of an existing sentence \cite{knight2002summarization, cohn2008sentence}. The constraint that the compressed sentence must be \emph{grammatical} results in substantial differences from our work. Much of the challenge stems from the need for supervised verbose-compressed pairs \cite{filippova2013overcoming} and the difficulty of incorporating unsupervised information \cite{turner2005supervised, miao2016language, fevry2018unsupervised}. Our work does not require that compressed sentences be grammatical, thus allowing us to consider unsupervised approaches.

There have been recent attempts to frame unsupervised summarization as an autoencoder problem \cite{miao2016language,fevry2018unsupervised,wang2018learning,baziotis2019seq}. However, most of these work uses continuous representations for the compressed form which does not fulfill our interpretability desideratum. \citet{west2019bottlesum} propose to summarize a sentence by extracting keywords from the sentence by maximizing the probability to predict next sentence. This approach is closely related to our approach, but differs in that it focuses on the global summarization rather than compressing each sentence independently.

Masked language models, such as the recently popular BERT~\cite{devlin2018BERT} and maskGAN \cite{fedus2018maskgan}, treat sentence generation as a series of masked token prediction problems. This is related to our task of identifying a longer sentence which originally generated a keyword sequence, but differs in that it uses explicit knowledge of the number and location of blanks. None of these approaches study the communication efficiency or compression rate tradeoffs implied by the masking and unmasking tasks.

Some work attempts to compress a sentence into a vector, such as the sentence variational autoencoder \citep{bowman2016continuous}.
Their focus is to learn embeddings, whereas we are interested in the communication aspect,
where continuous representations are not interpretable and not a medium users can output.
There is also some work that learns a communication scheme in an unsupervised manner \cite{foerster2016learning,sukhbaatar2016learning,lazaridou2017multi,mordatch2018emergence}.
\citet{lazaridou2017multi} attempt to make the scheme more interpretable by putting humans in the loop,
whereas we enforce interpretability structurally by restricting the scheme to keywords.

\paragraph{Applications.}
Autocomplete systems have been widely used for email autocomplete such as Gmail's Smart Compose and code autocomplete such as Pythia~\cite{svyatkovskiy2019pythia} and Deep TabNine.\footnote{\url{https://tabnine.com}} However, most of these systems focus on the left-to-right autocomplete setting where inputs are always the leftmost tokens in an input sequence, and the task is to predict the next token(s). This setting is inherently interpretable (much like the keyword constraint) but does not serve as an efficient communication mechanism, since a prefix of tokens rarely specifies the full meaning of a sentence.

Some work considers the keyword-based autocomplete setting in programming \cite{little2007keyword,miller2008inky}, but resorts to heuristics to restrict the search space and uses a hand-engineered scoring function to match keywords to target sentences. Our work differs in that it does not require experts to define available search space or design a matching algorithm specific to each domain.

\section{Illustrative example of generating keywords}
\label{sec:appendix_encoder}

\begin{figure}[h!]
\begin{center}
\setlength{\tabcolsep}{3.5pt}
    \begin{small}
    \begin{tabular}{c}
        \textbf{Sentence:} \textft{I will be 10 minutes late.} \\
        \\[-5pt]
        \framebox{{
            $\!\begin{tabular}{lccccccccc}
        $x$     & \textft{\shift} & \textft{i}  & \textft{will}  & \textft{be}  & \textft{10}  & \textft{minutes}  & \textft{late}  & \textft{.} \\
        \\[-5pt]
        $p$     & 0       & .1  & .2  & 0  & .9  & .7  & .8  & 0 \\
        $m$     & 0       & 0   & 0   & 0  & 1   & 1   & 1   & 0 \\ 
        \\[-5pt]
        $z$     & & & & & \textft{10} & \textft{minutes} & \textft{late} & & \\
            \end{tabular}$
        }} \\
        \\[-5pt]
        \textbf{Keywords:} \textft{10 minutes late} \\
    \end{tabular}
    \end{small}
\vskip -0.1in
\end{center}
\end{figure}

Given a target sentence, our encoder first tokenizes the sentence into a sequence of tokens $\x = (\x_1, ..., \x_m)$, predicts a probability $p_i$ for each token $\x_i$, and decides whether to keep each token ($m_i=1$) or not ($m_i=0$) by sampling from Bernoulli distributions according to the probability. The kept tokens become the keywords $\z = (\z_1, ..., \z_n)$.

\section{Details on experiments}
\label{sec:appendix_experiments}

\paragraph{Data.} We train all models on 500K sentences and test on 10K sentences of at most 16 tokens, randomly sampled from the Yelp restaurant reviews corpus~\cite{yelp2017yelp}. This dataset consists of many short, colloquial expressions that users may wish to autocomplete when writing reviews. We segment the reviews into sentences following the same procedure as \citet{guu2018edit} while treating capitalization as a separate, special token (e.g. \textft{I} $\rightarrow$ \textft{$<$shift$>$ i}) to emulate the keystrokes necessary for typing. 

\paragraph{Choice of evaluation metrics.} We use exact match as our accuracy metric. Other metrics such as the mean reciprocal rank (MRR) and BLEU are not considered because the time for choosing from top $n$ suggestions or revising wrong suggestions is likely to incur a large variance in the experiment results. In the user study, we additionally measure accuracy up to paraphrasing using user feedback.

\paragraph{Implementation.} For our encoder, we jointly train 300-dimensional word embedding and encode a sequence of tokens (i.e. target sentence) using a uni-directional LSTM with 300 hidden units. Then, we use a linear layer to output a probability of keeping each token and sample from Bernoulli distributions according to these probabilities. Finally, we output a sequence of kept tokens as keywords.

For our decoder, we use a standard sequence-to-sequence model~\cite{sutskever2014sequence}~with global attention~\cite{luong2015translation}~and copy mechanisms~\cite{see2017point}. Concretely, we jointly train 300-dimensional word embedding and encode a sequence of tokens (i.e. keywords) using a bi-directional LSTM with 300 hidden units. We use another uni-directional LSTM with 300 hidden units, which takes the embedding of each token at each time step, concatenated by the last hidden state of the LSTM, to either generate next token from a fixed vocabulary or copy one of the input tokens. 

All parameters are initialized by sampling from a uniform distribution between $-0.1$ and $0.1$. For optimization, we use the Adam optimizer \cite{kingma2014adam} with a learning rate of 0.001 for the encoder and decoder and 0.01 for $\lambda$, a minibatch size of 128, and the initial value of $5$ for $\lambda$.

\begin{table*}[t!]
  \begin{center}
  \small
\setlength{\tabcolsep}{2.5pt}
    \begin{tabular}{lr|rrr|rrrr}
      \toprule
                 &          & \multicolumn{7}{c}{Decoder (ACC@1 (\%))} \\
Encoder          & Kept (\%)  & \uniform(0.1) & \uniform(0.3) & \uniform(0.5) & \constrain(0.8) & \constrain(0.6) & \constrain(0.4) & \constrain(0.2) \\
      \midrule

\uniform~($\delta$=0.1)      &     21.34&      0.65&      0.54&      0.71&      0.23&      0.57&       0.70&      0.68\\
\uniform~($\delta$=0.3)      &     38.85&      2.68&      2.88&      3.56&      1.89&      2.37&      2.51&      3.68\\
\uniform~($\delta$=0.5)      &     56.71&      6.52&      9.19&     10.33&      7.08&      7.27&      6.84&      9.07\\
\hline
\constrain~($\eps$=0.8) &     25.48&      1.05&      1.29&      1.42&      1.15&      1.56&       2.00&      2.18\\

\constrain~($\eps$=0.6) &    34.04&       2.30&      2.58&      3.34&      3.43&      7.14&      8.13&      4.31\\
\constrain~($\eps$=0.4) &    41.87&      3.76&      4.55&      6.14&       8.70&     14.45&     13.13&      7.66\\
\constrain~($\eps$=0.2) &     55.80&     14.04&     18.88&     22.04&     38.31&     27.18&     22.81&      23.90\\

      \bottomrule
    \end{tabular}
  \end{center}
  \caption{\label{tab:robustness} We measure robustness as the performance of trained decoders with respect to different encoders. Selected uniform and constrained models have similar retention rates. The diagonal numbers represent the accuracy of matched encoder-decoder pairs.}
  \vspace{-1em}
\end{table*}

\section{Robustness of learned schemes}
\label{sec:appendix_robustness}
We quantify the robustness of communication schemes by measuring how well a decoder that was trained with respect to one encoder works with another encoder. 
From \tabref{robustness}, we first observe the followings as sanity checks: our trained encoders select better keywords than uniform encoders. For instance, using keywords from the trained encoder with a 55\% retention rate ($\text{\constrain}(0.2)$) improves the performance of the \uniform~decoders by 200\% compared to the matched uniform random encoder ($\text{\uniform}(0.5)$).
Furthermore, trained decoders are robust to being given more tokens than they were trained on (e.g. going from $\text{\uniform}(0.1)$ to $\text{\uniform}(0.5)$ encoders and $\text{\constrain}(0.8)$ to $\text{\constrain}(0.2)$ encoders), which ensures that users can increase the accuracy by passing more tokens.

We find that using loose constraints on reconstruction loss (i.e. large $\eps$) improves robustness to mismatched encoders. For example, even though the matched encoder-decoder pair of $\text{\constrain}(0.8)$ gets much lower accuracy (1.15\%) than that of $\text{\constrain}(0.2)$ (23.90\%), the $\text{\constrain}(0.8)$ decoder achieves even higher accuracy (38.31\%) than the matched encoder-decoder pair for $\text{\constrain}(0.2)$ (55.80\%). 
In the user study, we used the  $\text{\constrain}(0.6)$~model to balance decoder performance with performance under matching encoders.

\section{Analysis on learned schemes}
\label{sec:appendix_analysis}

\begin{figure}[h!]
\centering
\begin{minipage}{.45\textwidth}
  \centering
  \includegraphics[width=1\linewidth]{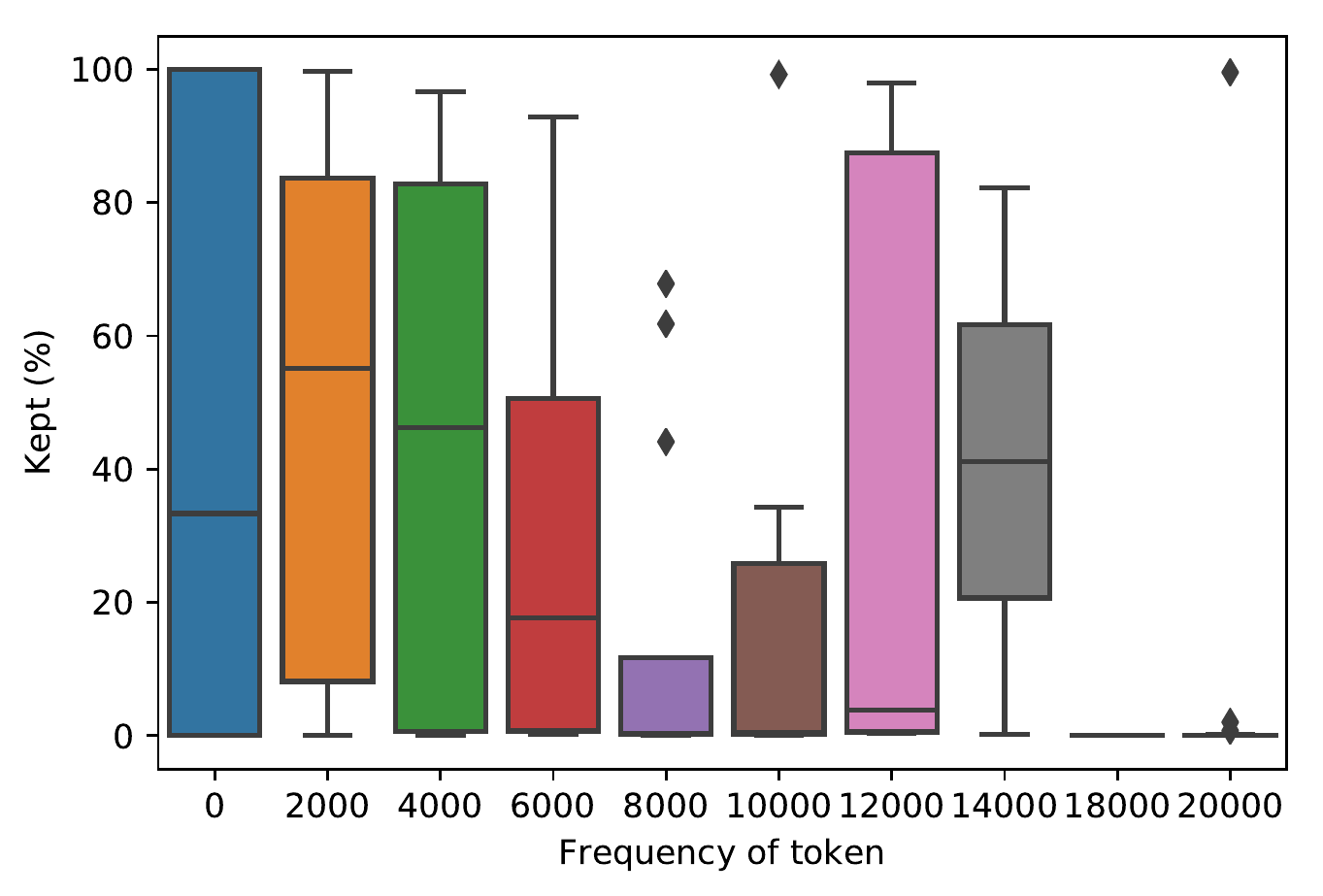}
  \captionof{figure}{Correlation between the frequency and retention rates of tokens.}
  \label{fig:freq}
\end{minipage}
\hfill
\begin{minipage}{.45\textwidth}
  \centering
    \begin{tabular}{llr}
      \toprule
        POS          & Examples      & Kept (\%) \\
      \midrule
        Determiner   & the, a, this & 3.96      \\
        Conjunction  & and, but     & 4.85      \\
        Pronoun      & it, you, we  & 9.79      \\
        Interjection & oh, wow      & 11.98     \\
        Verb         & love, recommend & 27.69  \\
        Preposition  & by, with     & 29.77     \\
        Adverb       & very, pretty & 34.99     \\
        Adjective    & delicious    & 35.92     \\
        Noun         & service, food & 40.04    \\
      \bottomrule
    \end{tabular}
  \captionof{table}{Part-of-speech tags in English and their percentages of being kept by our model.}
  \label{tab:pos}
\end{minipage}
\end{figure}

\paragraph{Correlation with frequency.} It is possible that the model simply drops frequent tokens and retains rare ones. We find that this is not the case as shown in \figref{freq}. While extremely frequent tokens are almost always dropped, tokens with moderate to low frequency have a wide range of behavior in terms of their retention rates. The figure indicates that there is not a strong correlation between the unigram frequency of a token on the x-axis and its corresponding retention rate from our model on the y-axis. 

\paragraph{Correlation with part-of-speech.} On the other hand, part-of-speech tags provide a stronger correlation to retention as shown in \tabref{pos}, with different part-of-speech tag categories resulting in large differences in retention rates. We broadly find higher retention rates for content words, and lower retention for determiners, conjunctions and pronouns. This observation matches our intuition of which tokens must be retained to capture the meaning of sentences.

\paragraph{Stability.} Finally, we observe that random restarts of our procedure can learn different communication schemes. However, even if the learned schemes differ in terms of their retention rates for individual tokens, they achieve similar overall retention and accuracy across restarts, suggesting that there are many distinct near-optimal schemes. For example, two random runs of our procedure achieve essentially identical overall retention rates (32.08\% and 33.52\%) and accuracies (19.41\% and 19.21\%), but the average difference of token-level retention rates weighted by frequency is 11.35\%. 

\section{Details on user study}
\label{sec:appendix_userstudy}

\paragraph{Data.} We sampled 500 sentences from a held-out set from the Yelp corpus restricted to sentences that appear more than once. Each user was randomly assigned 50 sentences from the pool to type.

\paragraph{Filtered responses.} We discarded 15 user responses where more than 20\% of keywords are longer than target sentences (autocomplete task), and 9 user responses where more than 20\% of paraphrased sentences are shorter than half of target sentences (writing task). Lastly, we filtered parts of user responses with exceptionally long time elapse (1.5 standard deviations above the mean, which is equivalent to $1.35$ minutes for keywords and $1.43$ minutes for sentences) to account for idle time.

\section{Interface for user study}
\label{sec:appendix_ui}

\begin{figure}[h!]
\centering
\vspace{-1em}
\begin{minipage}{.48\textwidth}
  \centering
  \includegraphics[width=1\textwidth]{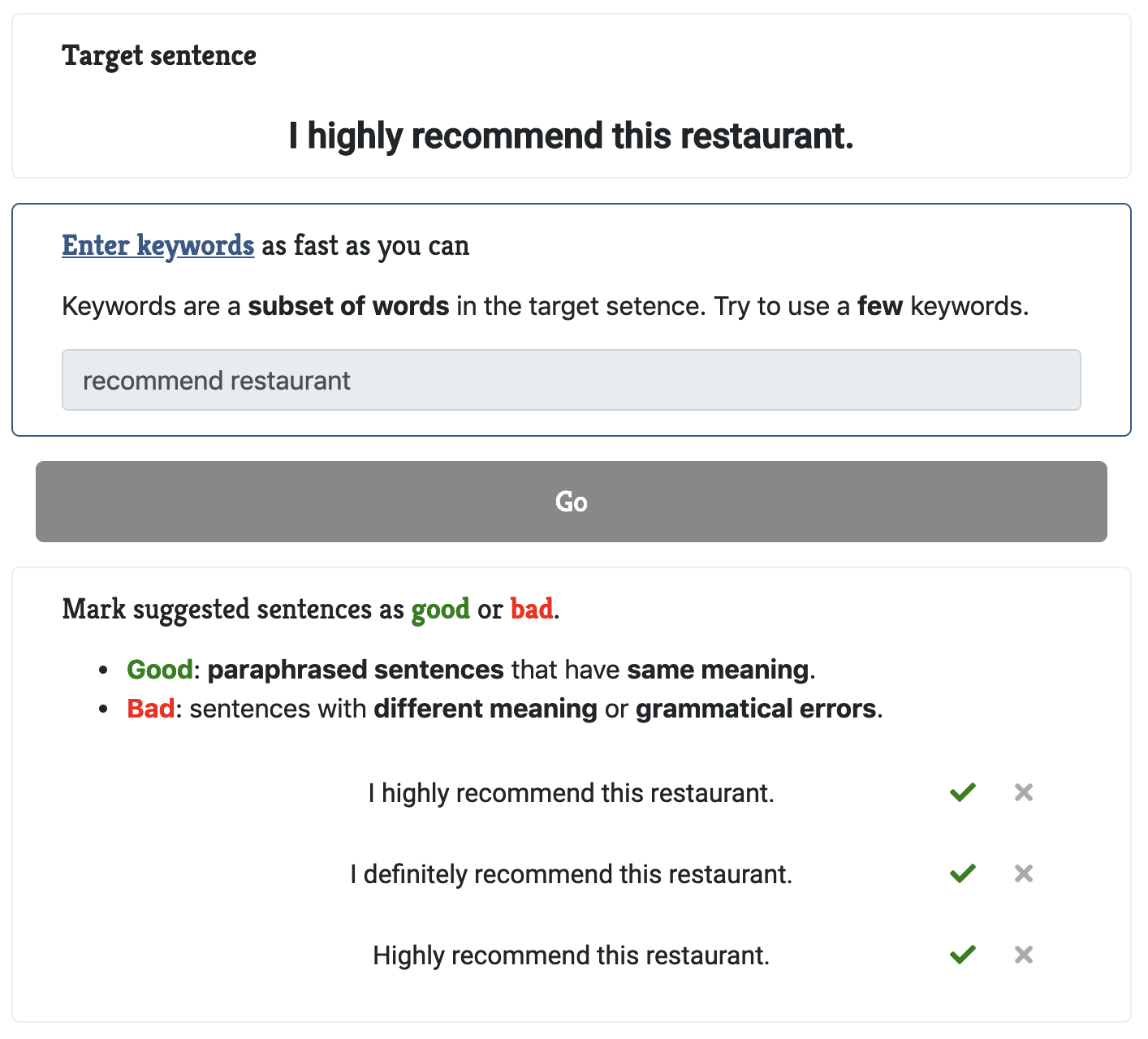}
  \caption{Autocomplete task interface.} 
\end{minipage}
\hfill
\begin{minipage}{.48\textwidth}
  \centering
  \includegraphics[width=1\textwidth]{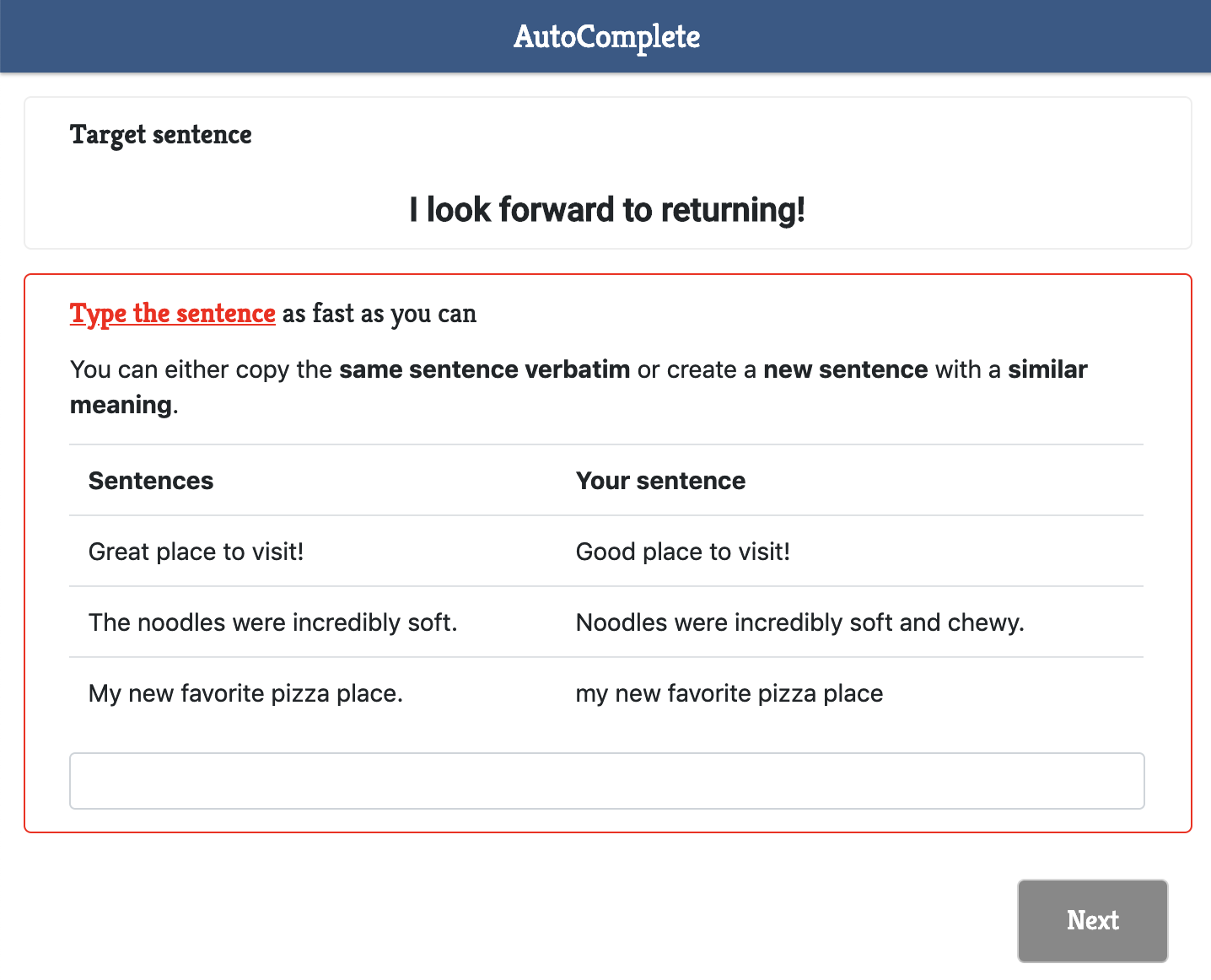}
	\caption{Writing task interface.}
\end{minipage}
\end{figure}

\end{document}